\documentclass[10pt,twocolumn,letterpaper]{article}
\usepackage{cvpr}
\usepackage{graphicx}
\usepackage{amsmath}
\usepackage{amssymb}
\usepackage{booktabs}
\usepackage{comment}
\usepackage{multirow}
\usepackage{verbatim}
\usepackage[dvipsnames]{xcolor}
\usepackage{shortbold}
\usepackage{float}
\usepackage{tikz}
\usepackage{color}
\usepackage{cite}
\usepackage{booktabs}
\usepackage[export]{adjustbox}
\usepackage{colortbl,hhline}

\usepackage{overpic}
\usepackage{enumitem} %
\usepackage{overpic} %
\usepackage{color}
\usepackage{xspace}

\definecolor{turquoise}{cmyk}{0.65,0,0.1,0.3}
\definecolor{purple}{rgb}{0.65,0,0.65}
\definecolor{dark_green}{rgb}{0, 0.5, 0}
\definecolor{red}{rgb}{0.8, 0.2, 0.2}
\definecolor{darkred}{rgb}{0.6, 0.1, 0.05}
\definecolor{blueish}{rgb}{0.0, 0.3, .6}
\definecolor{light_gray}{rgb}{0.7, 0.7, .7}
\definecolor{pink}{rgb}{1, 0, 1}
\definecolor{greyblue}{rgb}{0.25, 0.25, 1}
\definecolor{msorange}{rgb}{0.93,0.49,0.19}
\definecolor{msblue}{rgb}{0.33,0.56,0.76}

\definecolor{darkyellow}{rgb}{0.796,0.901,0}

\newcommand{\bfpar}[1]{{\vspace{0mm} \par \noindent \bf{{#1}}}}

\newcommand{\parnobf}[1]{\bfpar{#1.}}

\newcommand{\method}{3DFIRES\xspace}

\definecolor{BlueCam}{rgb}{0.10196, 0.52157, 1.0}
\definecolor{GreenCam}{rgb}{0, 0.5, 0}
\definecolor{PurpleCam}{rgb}{0.74,0.415,0.831}

\newcommand{\Fig}[1]{Fig.~\ref{fig:#1}}

\newcommand{\Tab}[1]{Tab.~\ref{tab:#1}}

\newcommand{\Sec}[1]{Sec.~\ref{sec:#1}}

\usepackage{blindtext}

\renewcommand{\paragraph}[1]{\vspace{1em}\noindent\textbf{#1}.}

\makeatletter
\DeclareRobustCommand\onedot{\futurelet\@let@token\@onedot}
\def\@onedot{\ifx\@let@token.\else.\null\fi\xspace}

\def\eg{\emph{e.g}\onedot}

\def\etc{\emph{etc}\onedot} 
\def\wrt{w.r.t\onedot} 
 
\def\etal{\emph{et al}\onedot}
\makeatother

\usepackage[accsupp]{axessibility}
\definecolor{cvprblue}{rgb}{0.21,0.49,0.74}
\usepackage[pagebackref,breaklinks,colorlinks,citecolor=cvprblue]{hyperref}
\def\paperID{1628}
\def\confName{CVPR}
\def\confYear{2024}
\newcommand{\ddr}{d_\textrm{DR}}
\newcommand{\rayr}{\vec{\rB}}
\begin{document}
\title{3DFIRES: Few Image 3D REconstruction for Scenes with Hidden Surfaces}
\author{Linyi Jin$^1$, Nilesh Kulkarni$^1$, David F. Fouhey$^2$ \\
University of Michigan$^1$, New York University$^2$ \\
{\tt\small \{jinlinyi,nileshk\}@umich.edu, david.fouhey@nyu.edu}}

\maketitle

\begin{abstract}
This paper introduces \method, a novel system for scene-level 3D reconstruction from posed images. Designed to work with as few as one view, \method reconstructs the complete geometry of unseen scenes, including hidden surfaces. With multiple view inputs, our method produces  full reconstruction within all camera frustums. A key feature of our approach is the fusion of multi-view information at the feature level, enabling the production of coherent and comprehensive 3D reconstruction. 
We train our system on non-watertight scans from large-scale real scene dataset. We show it matches the efficacy of single-view reconstruction methods with only one input and surpasses existing techniques in both quantitative and qualitative measures for sparse-view 3D reconstruction. Project
page: \url{https://jinlinyi.github.io/3DFIRES/}

\end{abstract}

\section{Introduction}
\begin{figure}[t]
\centering
\includegraphics[width=\linewidth]{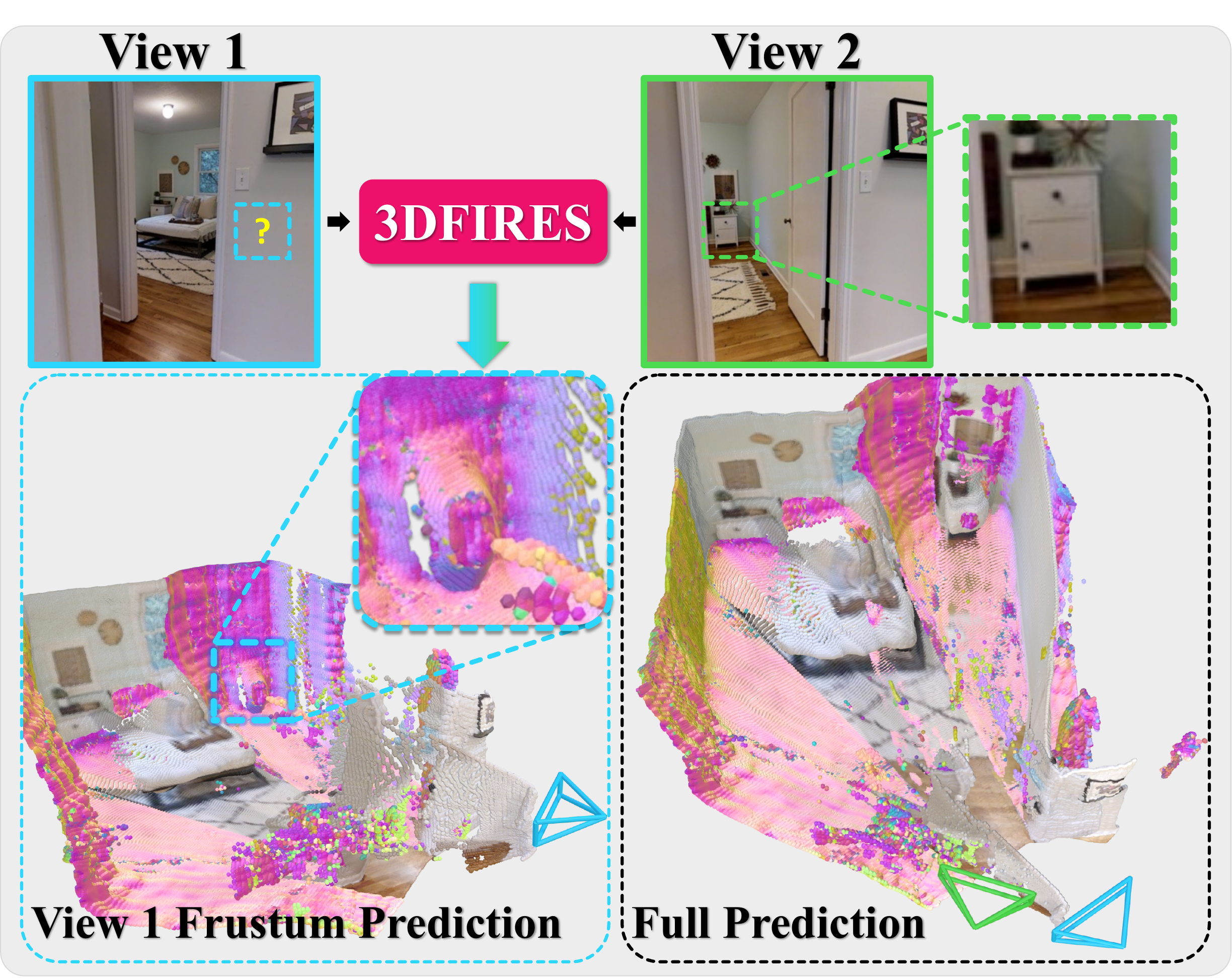}  \\
\caption{{\bf Reconstructing 3D from sparsely posed images}. Given a sparse set of posed image views, our method is able to reconstruct the full 3D of the scene. On the top, we show two sparse views of the scene in \textcolor{BlueCam}{View 1}  
 and \textcolor{GreenCam}{View 2}. On the bottom left is the 3D reconstruction from our network in the frustum of \textcolor{BlueCam}{View 1}. We show that our method can generate the occluded side table (zoom in). On the bottom right is the full reconstruction. We color occluded surfaces with surface normals.  }
\label{fig:teaser}
\vspace{-1em}
\end{figure}

Consider two views of the scene in~\Fig{teaser}. Part of the bedroom in \textcolor{BlueCam}{View 1} is occluded by the wall, and so you may be uncertain what is behind it, although you might guess the wall continues. Now consider adding in \textcolor{GreenCam}{View 2}. You can see a bedside table, but little else. However, you can fuse these pieces together to create a consistent 3D sense of the scene viewed by the images, including both the visible and invisible parts. We use this sense when shopping for real estate or looking at a friend's photos. 
We estimate the structure of the scene from parts that are visible to all views; integrate information across images for parts that visible in one view but not others; and take educated guesses for completely occluded regions. Importantly, as the available data increases from one camera to a handful, we can seamlessly integrate the evidence across views.

This task poses a challenge for current computer vision since it requires making judgments about visible and occluded 3D structures and integrating information across images with large pose change. These abilities are usually independently investigated in two separate strands of research.
With single image reconstruction techniques~\cite{wu2023mcc,kulkarni2022directed,tulsiani2018factoring,park2019deepsdf, Izadinia_2017_CVPR}, one can predict both visible and occluded 3D structure from an image, but stacking such outputs from multiple images can produce inconsistent outputs.
When handled independently, methods cannot identify the best view to reason about an occluded region. 
Non-line-of-sight imaging involves transmitting and receiving signals to reveal hidden scenes, incompatible with standard camera images~\cite{Isogawa_2020_ECCV}.
Sparse view reconstruction methods ~\cite{jin2021planar, agarwala2022planeformers, tan2023nopesac} can create consistent reconstructions from two views; however, these approaches are limited to the visible parts of the scene that can decomposed into planes. Moreover, these methods are usually specialized to a particular number of images that can be accepted.

Recently, there has been considerable progress in generalized radiance fields, which produce full 3D representations. This occupancy representation and per-scene optimization has shown promising results by optimizing for novel view synthesis on single scenes from posed images sets~\cite{mildenhall2020nerf,sitzmann2020implicit, tancik2020fourfeat, deng2022depth}. Extending this line of work, methods like~\cite{yu2021pixelnerf, srt22} have shown an ability to predict novel views for unseen scenes from a few images. However, since these methods optimize for perceptual quality, the underlying geometry often has artifacts. Like them we also require one or more image views at input, but instead we predict an implicit function~\cite{kulkarni2022directed} that can reliably reconstruct both visible and occluded parts of previously unseen scenes.

We propose \method, {\bf F}ew {\bf I}mage 3D-{\bf RE}construction of {\bf S}cenes, which integrates information from a variable number of images to produce a {\it full reconstruction} of the scene. \method integrates information in the features space across a varying number of images, enabling it to identify how to best use the available image data to produce an accurate reconstruction at a point. As output, \method produces a pixel-aligned implicit field based on a generalization of the Directed Ray Distance Function~\cite{kulkarni2022directed,kulkarni2023d2drdf}, which enables high quality reconstructions. Thanks to integration in feature space, the results are more consistent than handling images independently: this is what enables reconstructing the bed-side table in ~\Fig{teaser}, even though it is hidden by the wall in one image. We found and document several design decisions in terms of training and network architecture needed to produce these results.

We evaluate our method on complex interior scenes  from Omnidata~\cite{eftekhar2021omnidata, scharstein2002taxonomy} dataset collected with a real scanner. We compare \method with the point-space fusion of state-of-the-art methods for scene-level full 3D reconstruction methods from a single image~\cite{wu2023mcc, kulkarni2023d2drdf}. Our experiments show several key results. First, \method produces more accurate results compared to existing works. The improvements are larger in hidden regions, and especially substantial when measuring consistency of prediction from multiple views. Second, ablative analysis reveals the key design decisions responsible for \method's success. Third, \method can generalize to variable views: we train on 1, 2, and 3 views and generalize to 5 views. Finally, \method can reconstruct when given LoFTR~\cite{sun2021loftr} estimated poses with known translation scale.

\section{Related Works}
We aim to produce a coherent 3D scene reconstruction given a single or a few images with wide baselines.

\parnobf{3D from Single Image}
Predicting a complete 3D scene from a single image is inherently ambiguous. 
Recently different 3D representations have been proposed to reconstruct complete 3D scenes (including occluded surfaces) such as layered depth~\cite{shade1998layered}, voxels~\cite{choy20163d, girdhar2016learning, tulsiani2018factoring, kulkarni20193d}, planes~\cite{jiang2020peek}, pointclouds~\cite{fan2017point, wu2023mcc}, meshes~\cite{Nie_2020_CVPR, gkioxari2019mesh, usl2022}, or implicit representation for objects~\cite{park2019deepsdf, mescheder2019occupancy} and scenes~\cite{kulkarni2022directed, kulkarni2023d2drdf, sitzmann2020implicit, chibane2020ndf,dahnert2021panoptic}.  
While they have strong performance on single image, they do not necessarily produce coherent results when required to infer on multiple images of the same scene~\cite{kulkarni2023d2drdf}. Our method can reconstruct hidden geometry from at least a single image using implicit representation from~\cite{kulkarni2022directed}. Instead of naively fusing point clouds from different images, we fuse features when predicting a multi-view consistent point cloud with few input images.

\parnobf{3D from dense views}
Traditional multi-view 3D reconstruction methods can produce accurate and coherent pointclouds from pixel correspondences~\cite{scharstein2002taxonomy}. Classical methods in computer vision use approaches like Multi-view stereo (MVS) to construct only visible parts of the scene in all the images. There is a long line of work in trying to reconstruct scenes from video sequences~\cite{schonberger2016structure, davison2007monoslam} where they reconstruct visible scenes and camera poses. Learning-based methods for MVS estimate geometry for scenes~\cite{kar2017learning, sun2021neuralrecon, murez2020atlas, xie2022planarrecon} also require an input video to explicitly predict scene geometry. Instead of requiring high overlap inputs such as video frames, our method works on wide-baseline images.

\parnobf{3D from sparse view inputs}
Our approach operates in a multi-view setting with a sparse set of views. We have a similar setting as wide-baseline reconstruction~\cite{pritchett1998wide}. Associative3D~\cite{qian2020associative3d} reconstructs the whole scene but requires voxelized scenes to train, our method works on non-watertight scene data. Prior work also explores planar representation~\cite{jin2021planar, agarwala2022planeformers, tan2023nopesac} for coherent 3D surfaces in non-watertight scenes. They use feed-forward networks to predict visible 3D surfaces for each view and merge them using predicted correspondences. Our approach leverages an implicit representation that accommodates non-watertight data, enabling the reconstruction of both visible and occluded surfaces. We fuse deep features from multiple views to predict DRDF representation from Kulkarni \etal~\cite{kulkarni2022directed}, producing a coherent reconstruction.

\parnobf{Novel view synthesis}
NeRF~\cite{mildenhall2020nerf} and its extensions~\cite{wang2021ibrnet, yu2021pixelnerf, zhou2023sparsefusion} optimizes per-scene radiance fields for novel-view synthesis, this requires many views and test-time optimization.  Due to its occupancy-based representation, extracting geometry often requires thresholding the density function, which leads to cloudy geometry with sparse input views. Our method directly predicts geometry from unseen images without the need for test-time optimization.
PixelNerf~\cite{yu2021pixelnerf} or SRT~\cite{srt22} can generalize to new scenes but their objectives optimize for photometric losses.

\section{Method}

\begin{figure*}[t]
\vspace{-1em}
\centering
\includegraphics[width=1\textwidth]{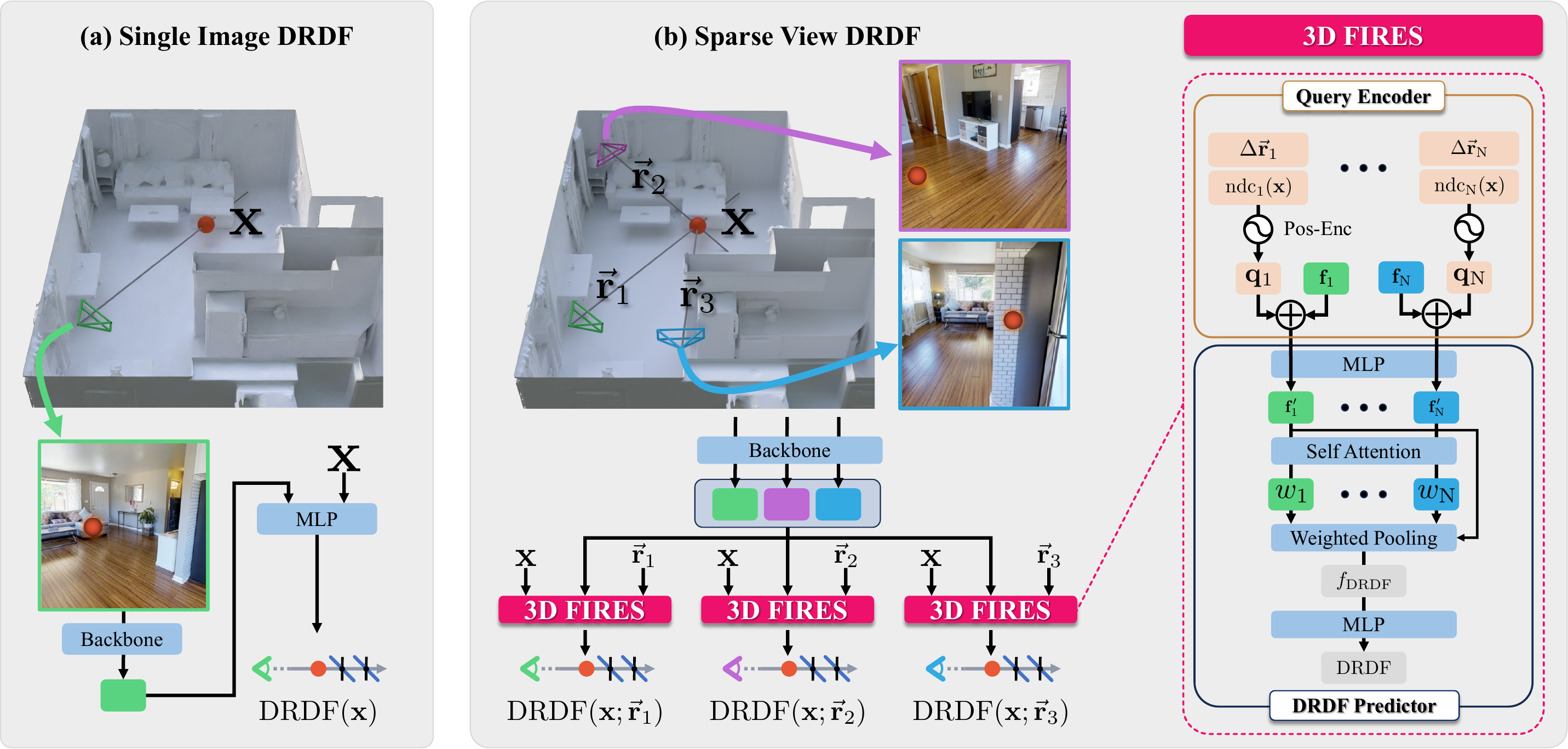}  \\
\caption{(a) Architecture for single view DRDF~\cite{kulkarni2022directed}. Given an image and a query pixel location, it predicts DRDF along the ray from the query pixel. (b) we extend (a) to work on sparse views. Middle: Given N images, a query point $\mathbf{x}$, and a query direction $\rayr_q$, we aggregate features from multiple images and output DRDF along the query ray. Right: We show detailed network architecture of \method which consists of a Query Encoder and a DRDF Predictor.}
\label{fig:method}
\vspace{-1em}
\end{figure*}

\begin{figure}[t]
\centering{
\includegraphics[width=0.48\textwidth]{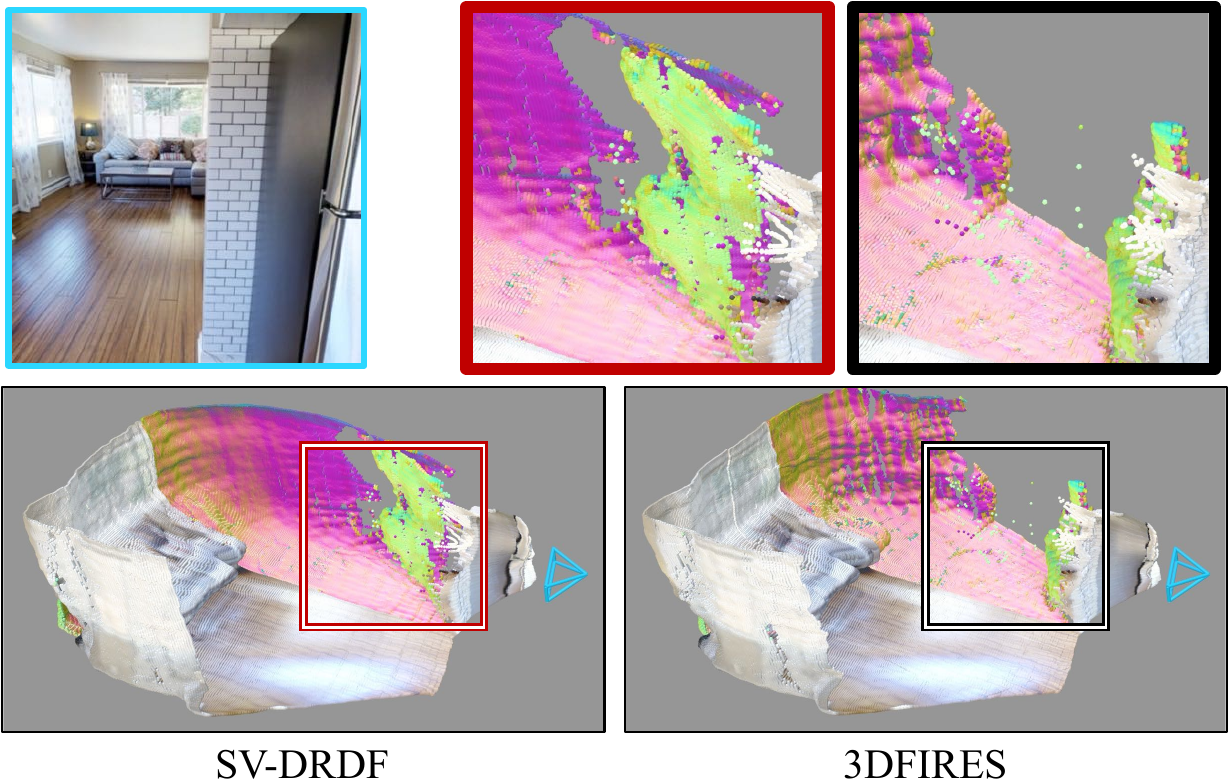}
}
\vspace{-1em}
\caption{{\bf Predictions in the \textcolor{BlueCam}{blue camera} frustum.} Occluded surfaces are colored with surface normals. A single image to 3D method like DRDF~\cite{kulkarni2022directed} is unable to reconstruct the parts of the scene behind the wall with certainty and hence erroneously adds a full wall in front of the hallway (red box). \method which fuses features from multiple views (\textcolor{GreenCam}{Green} and \textcolor{PurpleCam}{Purple} camera in \Fig{method}) predicts empty space for the entrance (black box).}

\label{fig:consistency}
\vspace{-2em}
\end{figure}

Our goal is to predict an accurate and consistent 3D reconstruction from one or more sparsely spaced camera views and known poses. With one image, the method should predict all surfaces in the camera frustum, including visible and occluded regions. With more images, the method should predict the surfaces in the union of the frustum. 

We tackle this problem with~\method, a simple and effective approach designed for this setting. We first discuss tackling scene reconstruction in a single image case in \S\ref{subsec:drdf} using the Directed Ray Distance Function (DRDF)~\cite{kulkarni2022directed} and scale this approach to multiple image views in \S\ref{subsec:mv_drdf}. In \S\ref{subsec:arch}, we show how we can operationalize our multi-view reconstruction goal with an attention-based model architecture.

\subsection{Background Single View Reconstruction}
\label{subsec:drdf}
We begin by revisiting the DRDF formulation for a single image reconstruction.
Consider a single image $\mathcal{I}$, a single view implicit reconstruction method aims to produce the full 3D reconstruction for the scene from this image. At inference, when conditioned on image features, the method outputs a distance function for a pre-defined set of 3D points in the camera frustum. It then decodes this predicted distance function to a surface to recover the 3D geometry of the scene. For instance, if the predicted 3D distance function is an unsigned distance function~\cite{chibane2020ndf}, the points on the surface are with distances close to zero.

Kulkarni \etal~\cite{kulkarni2022directed} solve the single image 3D reconstruction with the DRDF function and show that using the DRDF  outperforms the standard unsigned distance function. 
The DRDF is a ray-based distance function measuring the distance of a point $\xB$ to the nearest intersection with a surface along a ray $\rayr$. In
\cite{kulkarni2022directed}, the ray on which distances are measured is the ray from the camera center $\cB$ to $\xB$. 

\Fig{method} (a) shows the DRDF for one such ray. 
Now, any 3D point $\xB$ can be represented as its distance towards the camera times a unit ray direction, or $z\rayr$, where $z\in\mathbb{R}$ and $\rayr=\mathrm{norm}(\xB-\cB)$ where $\textrm{norm}(\pB) = \pB / ||\pB||$. The DRDF, $\ddr(z\rayr)$, furthermore includes a sign that determines for the point the direction along the ray towards the nearest intersection (i.e., forwards or backwards). Therefore $(z + \ddr(z\rayr) )\rayr$ corresponds to a point on the surface. 

The DRDF can be used to create a system that infers single image 3D by pairing the distance function at a point $\xB$ with pixel-aligned features. 
At inference time, as shown in \Fig{method} (a), given a point $\xB$ in the camera frustum we can extract corresponding pixel-aligned image features using an image backbone $\mathrm{BB}[\pi(\xB)]$, and use an MLP to predict the DRDF value corresponding to the point $\xB$ along the $\rayr$. Since DRDF is a ray-based function, its value only depends on the intersections along the ray. For any ray corresponding to a pixel on the image, the prediction of DRDF for the point depends on the image features, and the location of the point on the ray. This parameterization allows DRDF to learn sharp 3D reconstructions of the scene from a single RGB image. 
At training time, we train a model to predict the DRDF by supervising it with the ground-truth DRDF values computed using the mesh geometry.

\subsection{Extending DRDFs to Multiple Views}
\label{subsec:mv_drdf}
Now, with multiple views we have: N images $\{\mathcal{I}_{i}\}_{i=1}^{\text{N}}$, relative camera transforms $\{\pi_{i}\}_{i=1}^{\text{N}}$, and corresponding camera centers $\{\cB_i\}_{i=1}^{\text{N}}$, our goal is to reconstruct the 3D of the full scene.
While the task could perhaps be accomplished by simply predicting individual 3D for each camera, and assembling them together. Our insight is that if the camera frustums have considerable overlap, for overlapping regions we can achieve a better and more consistent reconstruction by allowing the network to reason about which camera provides the best view for each point.
This can be achieved by allowing the network to fuse features across cameras for the points in {\it feature} space rather than by concatenating in {\it point} space.
We propose to improve the feature quality of any point $\xB$ by fusing the features from multiple cameras. Since we are now dealing with the multi-view settings, a multi-view DRDF formulation is necessary to allow us to predict the DRDF value along each of the query rays, $\rayr_q$, originating from the respective camera centers.

In the case of multiple views, the image feature corresponding to a point $\xB$ should be a fusion of features $\{\fB_{\theta}[\pi_{i}(\xB)]\}_{i=1}^{\text{N}}$.
The feature should support predicting the N DRDF values along all the camera directions as $\{\ddr(z_{i}\rayr_{i})\}_{i=1}^{\text{N}}$. The intuition of our key idea is that multiple-image views provide more information about the 3D scene and hence potentially better features. We can learn these better features by fusing features to predict a consistent output. This requires a novel architecture that attends to features and rays, $\{\rayr_{i}\}_{i=1}^{\text{N}}$, originating from all the available image views. Under this formulation single view DRDF is a special case of our formulation where N is 1.

\subsection{Network Architecture}
\label{subsec:arch}
Towards the goal of predicting DRDFs along multiple query rays $\rayr_{q} \in \{\rayr_{i}\}_{i=1}^{\text{N}}$, we present a simple and effective network \method that accomplishes this task. \method consists of three modules: The first module is a \emph{Backbone Feature Extractor} that obtains pixel-aligned appearance features; by projecting the query point $\xB$ onto the camera, we can obtain a per-point and per-camera appearance feature as in~\cite{yu2021pixelnerf,kulkarni2022directed, wang2021ibrnet, saito2020pifuhd, mildenhall2020nerf}. Since the appearance feature is per-image, the model must learn to aggregate information across cameras. This is done with our second component \emph{Query Encoder} that provides geometric information for aggregating appearance features. Specifically, the query encoder uses the information about the relative positions of query point $\xB$ and query direction $\rayr_{q}$ \wrt cameras $\{\pi_i\}_{i=1}^{\mathrm{N}}$. 
The final module is the \emph{DRDF Predictor} that takes appearance and query features to produces a DRDF value along the query direction $\rayr_{q}$ by incorporating the appearance features (evidence for geometry) and query encoder features (evidence that relates different features).
\Fig{consistency} shows an example on how integrating information across multiple views leads to better prediction for occluded parts of the scene.

\parnobf{Backbone Feature Extractor} Our backbone features extractor $\text{BB}(\cdot)$ aims to create appearance features from an image. It accepts an image $\mathcal{I}_i \in \mathbb{R}^{H \times W \times 3}$ and produces a grid of D-dimensional features $\FB_i \in \mathbb{R}^{H' \times W' \times D_\text{img}}$. We use a  pre-trained depth estimating vision transformer~\cite{Ranftl2021}. Feature extraction for each image proceeds independently using the same network. With extracted per-camera backbone features, $\fB_{i}$, for point $\xB$ by interpolating features in $\{\FB_i\}_{i=1}^{\text{N}}$ at the projection $\{\pi_{i}(\xB)\}_{i=1}^{\text{N}}$ correspondingly.

\parnobf{Query Encoder} Our query encoder $q(\cdot)$ aims to enable a predictor to decide how to aggregate information across images. As input, the encoder takes a query 3D point $\xB$ and a query direction $\rayr_q$. It additionally considers the backbone features, camera centers $\{\cB_{i}\}_{i=1}^{\text{N}}$ and transforms  $\{\pi_{i}\}_{i=1}^{\text{N}}$.
Our query encoding is the concatenation of: (i) the relative viewing direction in camera $i$'s space $\Delta \rayr_{i}(\rayr_q)=[\rayr_q - \textrm{norm}(\xB-\cB_i), \rayr_q\cdot\textrm{norm}(\xB - \cB_i)  ]\in\mathbb{R}^4$; 
and (ii) the normalized device coordinates (NDC), coordinates of point $\xB$ in the camera frame $\textrm{ndc}_i(\xB)\in\mathbb{R}^3$. 
Intuitively this query representation, 
    $\qB_{i} = \{\Delta \rayr_{i}, \textrm{ndc}_i(\xB)\} \in \mathbb{R}^{7}$
enables reasoning such as: information about surfaces near $\xB$ in direction $\rayr_q$ is likely not visible in camera $i$ due to either angle or distance, so this feature ought to be weighted low. 
The ray query vector is encoded in a positional encoding layer~\cite{tancik2020fourfeat} with output dimension $D_{\text{query}}$.

\parnobf{DRDF Predictor} For a query ray and point tuple, $\{\rayr_q, \xB\}$, this model considers the image features $\{\fB_{i}\}_{i=1}^{\text{N}}$,  and query features $\{\qB_{i}\}_{i=1}^{N}$ yielding a joint camera specific feature, $\{\fB_{i}, \qB_{i}\}_{i=1}^{\text{N}}$, of dimension $D_{\text{img}}+D_{\text{query}}$. Our self-attention attends over all these features to produce a weight $w_{i}$ per feature. We aggregate the features using this weight to produce a fused feature for the point $\xB$. We then use the fused feature to predict a DRDF value between $[-1,1]$ with the help of an MLP. This is akin to selecting cameras that are likely to contain the geometry information about the ray point tuple and predicting the geometry information.

\subsection{Training \method}

The effectiveness of \method is improved by getting details right during training. One observation is that sampling points near intersections gives improvements over uniform sampling because the scene-level space is predominantly empty. By increasing the density of sampled points near surface, the network can better learn the scene structure. We sample points along the ray as per a Gaussian distribution centered at the intersection. Prior work~\cite{wang2021ibrnet} involves applying ray attention which allows for samples along a ray to attend with each other before the final prediction. This has been shown to be effective. 
However, combining ray attention with Gaussian sampling during training enables the network to `cheat'. Ray Attention exploits a train-time shortcut (query point density) to infer intersections. At inference as point density is uniform and this shortcut fails. Empirically we find Gaussian sampling alone to be more effective than ray attention.

\subsection{Implementation Details}
\bfpar{Training.} Our image feature backbone is vision transformer~\cite{Ranftl2021}  dpt\_beit\_large\_384 pretrained by MiDaS~\cite{Ranftl2022}. We use $\ell_1$ loss on log-space truncated DRDF~\cite{kulkarni2022directed, sun2021neuralrecon, dai2020sg}. During training, we randomly sample 1, 2, 3 views with $80$ rays per image and 512 points along each ray. Our method is trained for 300K iteration on NVIDIA A100 GPU with batch size of 1. More details in supp.
\bfpar{Inference.} Given N images, we extract backbone features
for each image. We generate $n_{\text{ray}}=128\times128$ query rays from each camera. Along each ray, we sample $n_{\text{pt}}=256$ points that have uniformly spaced depth from 0 to 8m. In total, we get N $\times n_{\text{ray}}\times n_{\text{pt}}$ query pairs $\{\xB, \rayr_q\}$, which are fed to \method in parallel to get DRDF value. We calculate positive-to-negative zero-crossings along each ray~\cite{kulkarni2022directed} to get a 3D point and aggregate the results.

\begin{figure*}[h]
\centering
\resizebox{\textwidth}{!}{
\begin{tabular}{c}
\includegraphics[width=\textwidth]{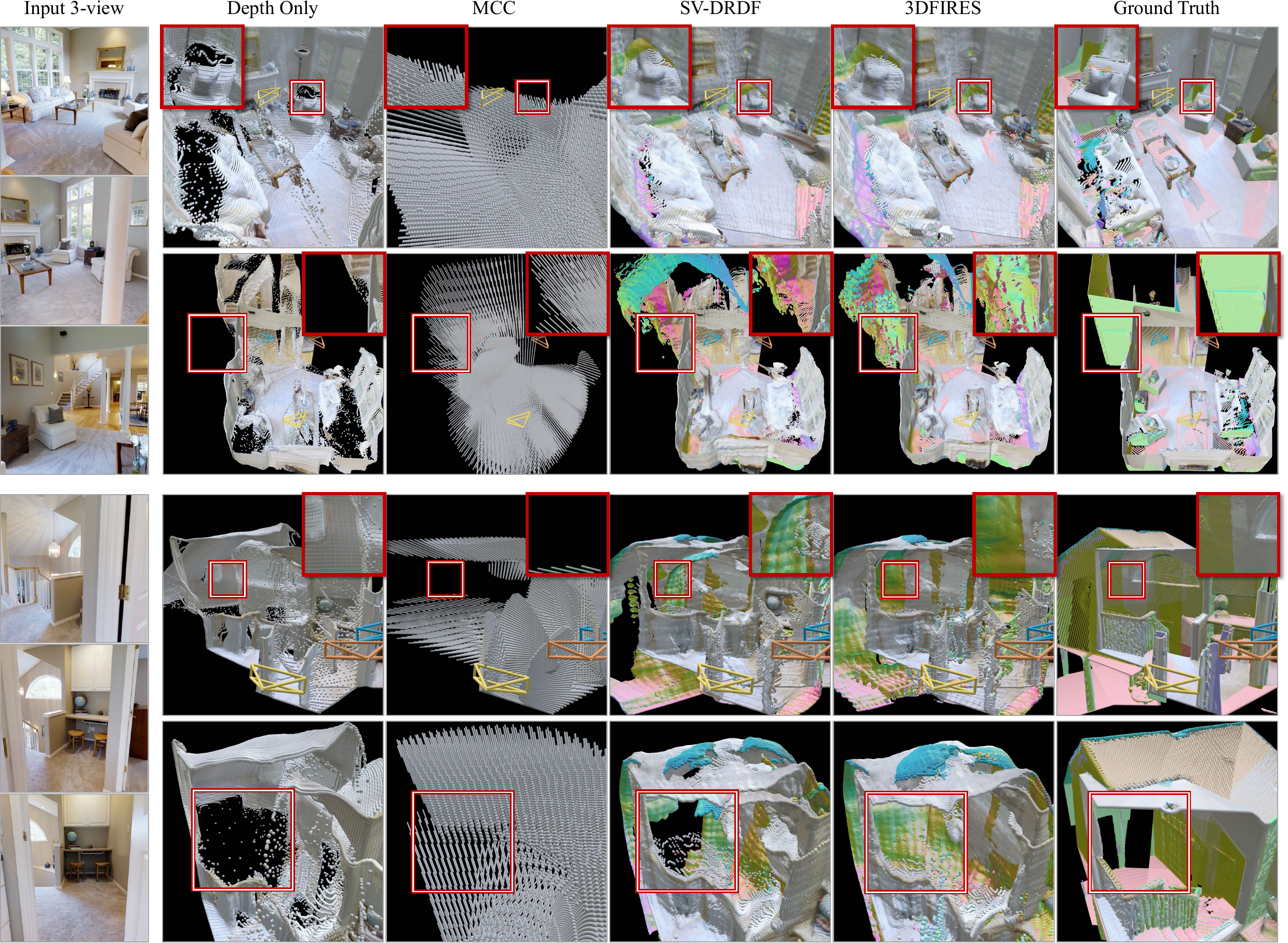}  \\
\end{tabular}
}
\caption{Comparison between different methods on held-out test scene. Occluded surfaces are colored with the computed surface normals. ``Depth only" leaves holes with sparse input views, \eg absent floors and walls. Occupancy-based method MCC~\cite{wu2023mcc} produces cloudy results, failing to get the details like pillow, tables. Concatenation of single view DRDF (SV-DRDF)~\cite{kulkarni2022directed} produces inconsistent results, \eg missing wall in row 2, the double wall in row 3. Our method produces more consistent predictions across different views and also recovers the hidden surface, resulting in a complete mesh. We urge the reader to see results provided in the supplementary videos.}
\label{fig:comparewall}
\vspace{-1em}
\end{figure*}

\section{Experiment}
In this section, we present the experimental framework for \method, our system designed to reconstruct full scene geometry from wide-baseline, sparse images. Considering the novelty of our problem, there is no prior work that does this exact setting.
To address this, we curated a dataset and developed testing metrics specifically tailored to the problem's requirements. 
We conduct comprehensive evaluations of \method using real scene images, comparing its performance against alternative methods in the field.

\subsection{Dataset}
Following~\cite{kulkarni2023d2drdf}, we use the dataset from the Gibson database~\cite{xia2018gibson}, which contains real images of complex and diverse scenes such as multi-floor villas and expansive warehouses. 
The scale of the assets in the dataset presents challenging reconstruction problem, which is desirable for evaluating the ability to recover occluded surfaces. 
We use the images sampled by Omnidata~\cite{eftekhar2021omnidata} for a diverse set of camera poses from the Taskonomy~\cite{Zamir2018CVPR} Medium subset, including 98/20/20 training/validation/test buildings.
Since our multiview setting is different from the single-view setting of~\cite{kulkarni2023d2drdf}, the precise samples are different. 
Our setting is also similar to~\cite{jin2021planar,tan2023nopesac} in that images have wide baselines (median 2.8m translation, 63.9$^\circ$ rotation), unlike methods using video frames~\cite{sun2021neuralrecon} where images have high overlap. Our approach diverges from~\cite{jin2021planar, tan2023nopesac} in also reconstructing occluded regions and using real (not rendered) images.

To curate our image sets, we use a sampling process like~\cite{jin2021planar}. For a set of $k$ images, after picking an image at random, each new image is selected to have at most 70\% overlap with any existing image in the set, and at least 30\% overlap with at least one other image in the set. The process balances diversity and coherence in the viewpoints. We crop images to a fixed field of view. We collect 3781 training sets among $\ge10$K images. We also sample 300 sets of 3-view images and 100 sets of 5-view images for evaluation from the held-out test scenes. 
See the supplementary for dataset generation details. The 3 view and 5 view test set contain considerable occluded 3D geometry (41.9\% and 43.7\% respectively).

\subsection{Baselines}
To the best of our knowledge, no prior work reconstructs occluded regions from sparse-view images at scene scale. 
We thus create strong baselines from existing methods that handle parts of our setting.
Each method is the strongest in its line of work. 

For instance, the visible surface upper-bound includes all methods that reconstruct visible surfaces from sparse views~\cite{tan2023nopesac,sun2021neuralrecon,jin2021planar}.
The DRDF method~\cite{kulkarni2022directed,kulkarni2023d2drdf} has been shown to be more effective for scene-level 3D reconstruction compared to many  other implicit functions like density~\cite{yu2021pixelnerf}, occupancy~\cite{saito2020pifuhd}, unsigned distance functions on scenes and rays~\cite{chibane2020ndf}. 
MCC~\cite{wu2023mcc} is likewise SOTA for point cloud completion.

\bfpar{Depth Only~\cite{Ranftl2021,eftekhar2021omnidata}} 
Prior state-of-the-art works on sparse scene reconstruction~\cite{tan2023nopesac,sun2021neuralrecon} predict visible surfaces from multiple views, but cannot recover hidden surfaces.
To show the near-oracle reconstruction of visible surfaces, we use MiDaS~\cite{Ranftl2021} depth model trained on Omnidata~\cite{eftekhar2021omnidata} {\it with ground-truth scale and shift}. This baseline is an upper bound on the performance of methods like~\cite{tan2023nopesac, jin2021planar, agarwala2022planeformers, sun2021neuralrecon}.

\bfpar{Multiview Compressive Coding (MCC)~\cite{wu2023mcc}} This method predicts  occupancy probability from RGB-D partial pointclouds. MCC works on scene-level reconstructions including non-watertight meshes. 
We train MCC on the same training set as ours. This method requires depth as input and at inference we provide it with ground truth depth. 
Since MCC only works on a single point cloud, to produce predictions from multiple images, we infer each image independently and aggregate the predicted point cloud in point cloud space. 

\bfpar{Single-view DRDF (SV-DRDF)~\cite{kulkarni2022directed}} This method reconstructs both visible and hidden surfaces from a single input image. 
We use this baseline to show the benefit of our proposed multi-view feature aggregation. 
For a fair comparison, we upgrade the original backbone from ResNet34~\cite{he2016deep} to the same BEiT~\cite{Ranftl2021} and use the same training strategy such as Gaussian sampling of points. Both improve results. 
Since this baseline only supports single image reconstruction, we produce predictions independently from each input image and aggregate all the point clouds.

\begin{figure*}[t]
\vspace{-2em}
\centering
    \begin{center}
    \includegraphics[width=\textwidth]{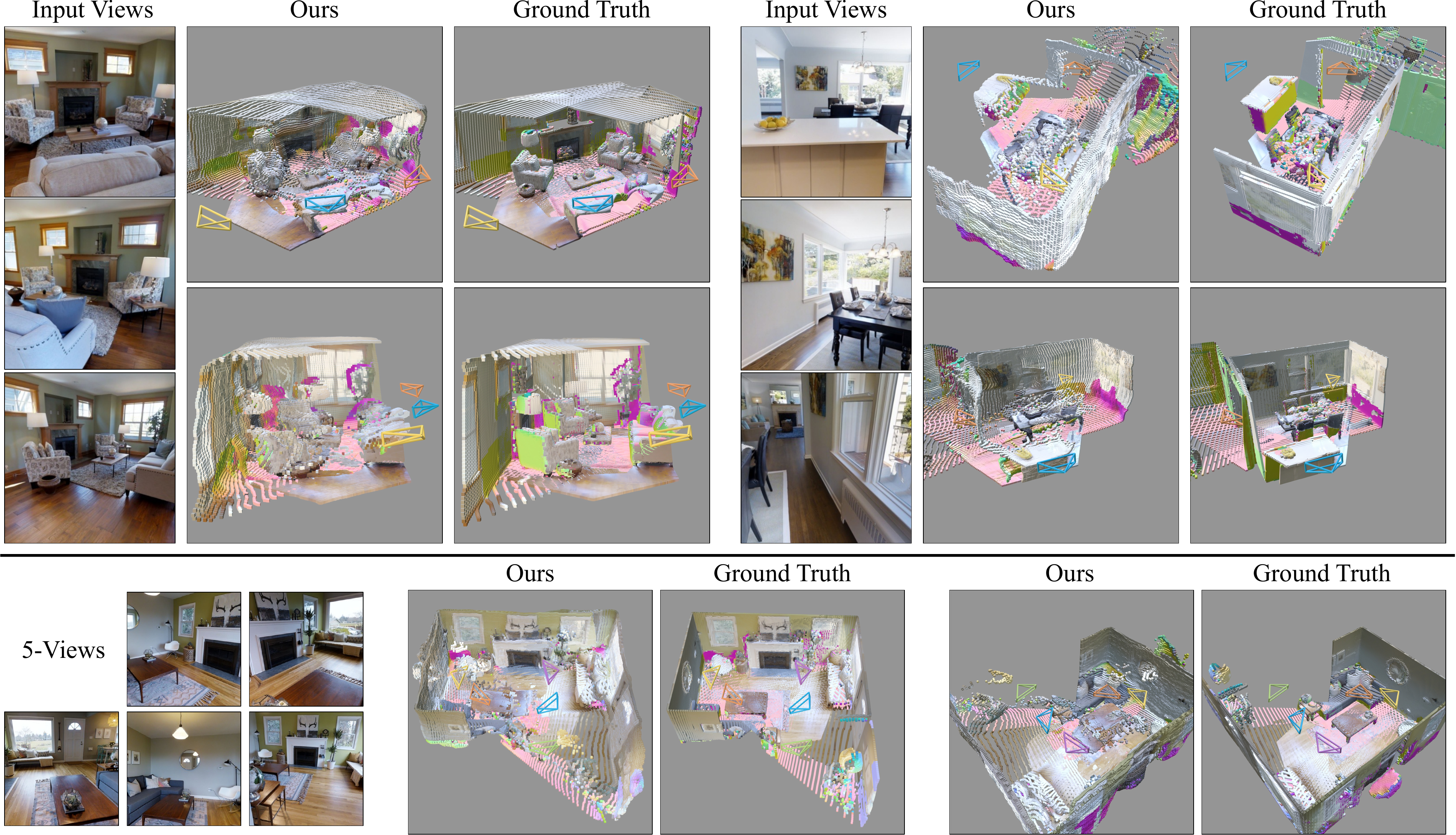} 
    \end{center}
    \vspace{-6mm}
    \caption{Qualitative results on held-out test scenes. Top row: Reconstruction from 3 images and compared with ground truth. Our method can reconstruct a complete scene structure within all the camera frustums, including the occluded surfaces. Bottom row: Predictions from 5 input images  compared with ground truth. For the 2nd and 3rd examples, ceilings are removed to reveal the details of the scene.
    }
    \label{fig:goodwall}
    \vspace{-1em}
\end{figure*}

\subsection{Evaluation Metrics}
We use two metrics to evaluate our system.
\bfpar{Scene F score.} Following~\cite{kulkarni2022directed, wu2023mcc}, we compute the scene accuracy (fraction of predicted points within $\rho$ of a ground truth point), completeness (fraction of ground truth points within $\rho$ from a predicted point), and their F-score (F1). 
This gives an overall summary of scene-level reconstruction. We classify the scene into (1) visible: points that are visible from any one of the input views; and (2) hidden: points that are hidden from all of the input views.
Due to the space limit, we only show F-score at $\rho=0.2$. A full table with accuracy, completeness, F-score at different $\rho$ is in the supp. Trends are the same across values of $\rho$ and there is no significant accuracy/completeness imbalance for the baselines (MCC, SV-DRDF).

\bfpar{Multiview consistency.}
Only measuring the F-score does not measure the consistency of 3D reconstruction when generating results from multiple views. Doubled predictions of surfaces do not change the Scene F score results if they are within $\rho$. 
Prior work~\cite{jin2021planar} used a detection-based method that penalized double surfaces on planar predictions, but their metric is not applicable since it requires planar instances. We require a metric that can measure the consistency of 3D reconstruction of points in individual frustums. Specifically, we would like to ensure that points $\PB_{i}$ generated from all query rays originating from $\cB_{i}$ of $\pi_{i}$ are consistent with points, $\PB_{j}$, generated from by ray queries from $\cB_{j}$ of $\pi_{j}$ at the intersection of frustums of both the cameras. For every point, $\pB \in \PB_{j}$ and within the field of view of camera $i$, we compute their minimum distance to points in $\PB_{i}$. Our metric measures percent of points in the set $\PB_{j}$ that have minimum distance within the threshold of $\rho$. We evaluate this metric bidirectionally to ensure complete results.

\subsection{Results}

\begin{table*}[t]
\vspace{-1em}
\centering
\caption{Quantitative results on Scene F-score ($\rho=0.2$) for Hidden points, Visible points, All points. For $3$ and $5$ views, we evaluate Consistency. Depth only: visible surface upperbound is separated to indicate it has oracle information. Despite accurate reconstructions on visible surfaces, these lines of work cannot recover hidden surfaces, causing low overall performance. 
With $1$ view, \method is comparable to single view DRDF. 
With more views, \method outperforms all the other baselines in F-score. There is large improvement in consistency metric compared to single view DRDF, showing that aggregating features produces a more coherent reconstruction.  Full tables showing accuracy and completeness are in the supplemental.
} %

\resizebox{\linewidth}{!}{ %
\begin{tabular}{@{}l|ccc|cccc|cccc@{}}
\toprule
\multicolumn{1}{c|}{~} 
& \multicolumn{3}{c|}{1 view} 
& \multicolumn{4}{c|}{3 views}
& \multicolumn{4}{c}{5 views}\\

& Hidden $\uparrow$ & Visible $\uparrow$ & All  $\uparrow$
& Hidden $\uparrow$ & Visible $\uparrow$ & All  $\uparrow$ & Consistency $\uparrow$ 
& Hidden $\uparrow$ & Visible $\uparrow$ & All  $\uparrow$ & Consistency $\uparrow$\\ 
\midrule
Depth only 
& - & \textbf{85.31} & 60.12 
& - & \textbf{87.84} & 63.90 & 72.79
& - & \textbf{91.29} & 69.40 & 72.57 \\
\hline
MCC
& 40.27 & 56.40 & 50.25
& 42.91 & 62.02 & 54.78 & 70.20
& 38.51 & 64.44 & 55.94 & 66.57 \\
SV-DRDF 
& \textbf{53.36} & 73.45 & 65.21
& 48.02 & 76.19 & 65.61 & 76.44
& 47.51 & 81.31 & 70.54 & 78.13 \\
\method 
& 53.34 & 74.29 & \textbf{65.71}
& \textbf{49.99} & 76.74 & \textbf{66.56} & \textbf{85.48}
& \textbf{49.52} & 81.74 & \textbf{71.41} & \textbf{85.92} \\
\bottomrule
\end{tabular}
} %

\label{tab:tab1}
\vspace{-1em}
\end{table*}

\begin{table}[t]
\centering
\caption{Ablation study on training strategies. GS: Gaussian sampling near intersection along the ray during training. Ray Attn: points along a query ray attend to each other.
} %

\resizebox{0.9\linewidth}{!}{ %
\begin{tabular}{l|cccc}
\toprule
&Hidden & Visible & All & Consistency \\
\midrule
-GS & 43.07 &  77.05 & 64.81 & 83.45 \\
+Ray Attn. -GS & 47.09 & \textbf{77.60} & 65.58 & 83.27 \\
+Ray Attn. +GS & 14.85 & 3.36 & 13.29 & 33.56 \\
Ours & \textbf{50.20} & 77.30 & \textbf{66.46} & \textbf{85.45} \\
\bottomrule
\end{tabular}
} %

\label{tab:ablation}
\end{table}

\bfpar{Qualitative Results.}
\Fig{consistency} shows reconstruction from using query rays from the \textcolor{BlueCam}{blue camera} in \Fig{method}. Occluded surfaces are colored with surface normals. DRDF~\cite{kulkarni2022directed} is unable to reconstruct the parts of the scene behind the wall with certainty and erroneously adds a full wall in front of the hallway. \method fuses features from multiple images (\textcolor{GreenCam}{Green} and \textcolor{PurpleCam}{Purple} camera in \Fig{method}) accurately predicts the empty space. 

\Fig{comparewall} shows results unseen test scenes, and compares reconstruction of baselines. Red box crop show highlighted differences and provide a zoomed-in view for detailed examination. 
Depth only (MiDaS with ground truth scale and shift) reconstructs {\it only} visible regions this leaves holes such as the missing surfaces behind chairs in Row 1; and absent floor sections in Row 4. 
MCC~\cite{wu2023mcc} tends to produce cloudy volumes and misses details like pillows and tables. 
Single-view DRDF (SV-DRDF) produces occluded regions and sharp surfaces but lacks consistency when aggregating results from multiple views.
This is noticeable in its inability to reconstruct the occluded wall in Row 2, the creation of a doubled ceiling in Row 3 due to occlusions.
\method, effectively merges observations from multiple images, resulting in sharp and accurate reconstructions of both visible and hidden surfaces. By fusing information across views in the feature space, our method overcomes the limitations of other approaches. This ensures comprehensive and consistent scene-level reconstruction from few sparse views.

In \Fig{goodwall} we show additional alongside the ground truth. 
\method successfully reconstructs large occluded areas, floors hidden by foreground objects (colored in pink), and unseen sides of objects such as the back of chairs in the first example and the kitchen islands in the second example. The reconstruction from multiple views demonstrates consistency and coherent surfaces in overlapping regions.

While our method is trained with up to three views, it seamlessly extends to five views. This adaptability stems from our architecture's inherent flexibility to the number of input views. With increasing views it predicts clean and coherent reconstructions within all the camera frustums.

\bfpar{Quantitative Results.} We evaluate our method on sets of 1, 3, 5 views respectively, as detailed in \Tab{tab1}. Our approach, designed for flexible input views, matches prior works in single-view scene reconstruction and achieves state-of-the-art results with multiple input views. In single-image cases, it is comparable to the single-view DRDF baseline.

For 3-view sets, our method outperforms MCC~\cite{wu2023mcc} or DRDF~\cite{kulkarni2023d2drdf}. Although MiDaS with ground truth scale and shift demonstrates optimal visible surface reconstruction, it falls short in overall scene reconstruction because of no reconstruction on occluded surfaces. 
When evaluated on scene consistency, \method shows a large absolute improvement of $>$ 9\%, over the second-best baseline, showing \method's ability to aggregate features across views to produce consistent results. 

The trend persists with 5-view inputs, where our method has the highest F score and consistency. Our method is not trained on 5-views subset but still remains robust to more input views enhancing the reconstruction quality in both visible and hidden surface reconstructions.

\begin{table}[t]
\centering
\caption{Quantitative results on noisy camera poses generated by LoFTR, evaluated on 3 view cases at $\rho=0.2$. \method assumes accurate pixel-aligned features but still produces more consistent reconstructions compared to not aggregating features.
} %

\resizebox{0.9\linewidth}{!}{ %
\begin{tabular}{l|cccc}
\toprule
3-View &Hidden & Visible & All & Consistency \\
\midrule
SV-DRDF & 37.39 & \textbf{62.71} & 52.93 & 57.65 \\
Ours & \textbf{38.85} & 62.40 & \textbf{53.19} & \textbf{65.71} \\
\bottomrule
\end{tabular}
} %

\label{tab:loftr_gt_scale}
\end{table}

\subsection{Ablations and Analysis}

\bfpar{Ablation study on training strategy.}
We conduct an ablation study (\Tab{ablation}) to investigate the effectiveness of different training strategies for our method. 
Without Gaussian sampling or ray attention (-GS), the method has degraded performance (-7\% in hidden F score). With ray attention only (+Ray Attn. -GS), the method is able to better reconstruct the hidden surface but is still worse than ours (-3\%). With both ray attention and Gaussian sampling (+Ray Attn. +GS), the network finds shortcut during training and does not work during testing. With Gaussian sampling strategy, our method performs the best.

\bfpar{Robustness with noisy camera poses.} Our method requires accurate camera poses to aggregate pixel-aligned features. This setting is challenging with sparse view data since camera estimation can be noisy. We test if the misalignment of image features caused by noisy camera projection matrices degrades our system. 
We use LoFTR~\cite{sun2021loftr} to estimate the camera rotation and translation angle and evaluate the reconstruction within all the camera frustums. Since LoFTR does not provide a translation scale, we use ground truth instead. \Tab{loftr_gt_scale} shows results on 3-view cases.
Our method still has significantly higher consistency over single view DRDF baseline. We provide an analysis with synthetic Gaussian camera noise in the supplementary.

\section{Conclusions}
We present \method, a scene-level 3D reconstruction method that requires only one or a few posed images of a scene. 
Our method takes in an arbitrary number of input views, fuses multi-view information in the features space and predicts DRDF given a 3D point and query direction. We train our method on a large-scale scene dataset and show its strong ability to reconstruct both visible and hidden surfaces coherently within all the camera frustums on challenging wide-baseline images. Currently, our methods requires pose input from off-the-shelf estimation methods, solving for 3D reconstruction and adapting the poses is a challenging next step and left to future work.
\vspace{0.5em}
\small{
\parnobf{Acknowledgments} 
Thanks to Mohamed Banani, Richard Higgins, Ziyang Chen for their helpful feedback. Thanks to UM ARC
for computing support. Toyota Research Institute provided funds to support this work.
}

\clearpage
{
    \small
    \bibliographystyle{ieeenat_fullname}
    \bibliography{macros,main}
}
\clearpage
\setcounter{page}{1}
\maketitlesupplementary
The supplemental material shows: details about the dataset (\Sec{supp:dataset}), detailed descriptions of implementation (\Sec{supp:implementation}), and additional results (\Sec{supp:results}). 

\section{Dataset Curation}
\label{sec:supp:dataset}
To effectively test the ability to reconstruct hidden surfaces, we need images that contain large occluded areas. 
Omnidata already provides posed images with corresponding statistics such as occlusion score, camera pitch, walkability score, objectness score \etc. We use 
these statistics to filter data better suited for our task.
We remove images within three criteria. 1) images mainly contain walls (do not contain foreground objects) by a simple criterion: occlusion boundaries less than 1\% of the image. The occlusion boundary is provided by the Omnidata~\cite{eftekhar2021omnidata}.  2) Images with large invalid areas ($\ge10\%$ of invalid depth value, \eg windows or mirrors) 3) Images looking upward to the ceiling (pitch $\ge0$).  We crop all images to fixed FoV of 63.4$^\circ$ following~\cite{jin2021planar}.
After cleaning the images, we sample image sets from these single images.
For a set of $k$ images, after picking an image at random, each new image is selected to have at most 70\% overlap with any existing image in the set, and at least 30\% overlap with at least one other image in the set. The process balances diversity and coherence in the viewpoints. We collect 3781 training sets among $\ge10$K images. We sample 300 sets of 3-view images and 100 sets of 5-view images for evaluation from the held-out test scenes.

\begin{figure*}[t]
\centering
\scriptsize
    \vspace{-1em}
\resizebox{\linewidth}{!}{
\includegraphics[width=0.8\textwidth]{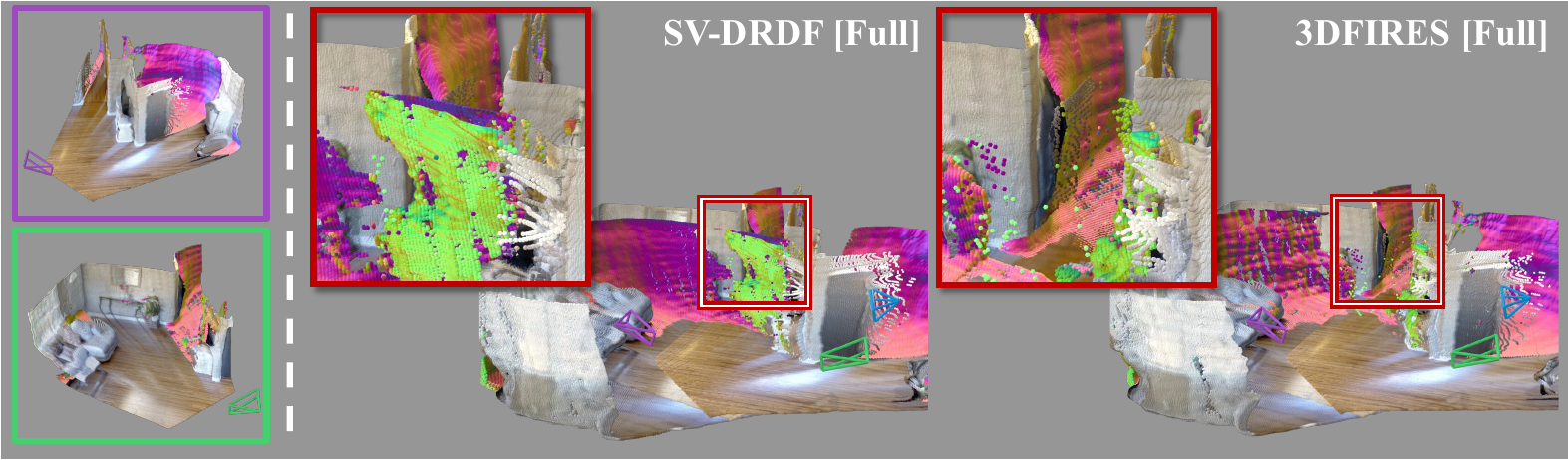}} \\
    \vspace{-1em}
    \caption{Results on single camera frustum of Fig. 3. 
    SV-DRDF concatenates the single view predictions, 
    so any inconsistency among the predictions affects the overall result.}
    \label{fig:supp:fig3}
\end{figure*}

\section{Implementation Details}
\label{sec:supp:implementation}
\bfpar{Network Architecture}
\Tab{supp:model_arch} shows a detailed network architecture.

\begin{table*}[t]
\caption{\textbf{Model Architecture}. Step (1) is the backbone feature extractor, which uses a ViT to extract image features from all input views. (2) given a query 3D points $\xB$, we get its pixel aligned feature through interpolation. (3) Our \textit{Query Encoder}, which enables a predictor to decide how to aggregate information across images. (4)-(6) is the \textit{DRDF predictor}, which concatenates the features in (4); self-attention attends all features across cameras to produce a weight per feature in (5); and uses the weight to produce a fused feature for the point $\xB$ and use a final linear layer to extract a 1D DRDF value in (6). }\label{tab:supp:model_arch}
\centering
\scriptsize
\resizebox{0.75\textwidth}{!}{
\begin{tabular}{@{}lllc@{}}
\toprule
Index & Inputs & Operation & Output Shape           \\ \midrule
(1)    & N Input Images & ViT Backbone & $N \times H \times W \times D_{\text{img}}$           \\
\hline
(2)   & query $\xB$, (1)   & Pixel aligned features $\{f_i\}_{i=1}^{\text{N}}$ & $N\times D_{\text{img}}$ \\
\hline
(3)   & query pair ($\xB$, $\rayr_q$)    & \begin{tabular}[c]{@{}l@{}}
\textit{Query Encoder $q(\cdot)$}:  \\ $\Delta\rayr_i\leftarrow\text{concat}[\rayr_q-\text{norm}(\xB-\cB_i), \rayr_q\cdot\text{norm}(\xB-\cB_i)]$,\\
$\qB_i\leftarrow$concat[PE($\Delta\rayr_i$), PE(ncd($\xB_i$))], \\\end{tabular} 
& $N\times D_{\text{query}}$
\\
\hline
(4)   & (2) (3)    & \begin{tabular}[c]{@{}l@{}}
concat($\qB_i, \fB_i)$,\\
Linear($D_{\text{img}}+D_{\text{query}}$, $D_{\text{feat}}$) 
\end{tabular}            
& $D_{\text{feat}}$\\
\hline
(5)   & (4)   & 
\begin{tabular}[c]{@{}l@{}}
Linear($D_{\text{feat}}$, $D_{\text{feat}}$)\\
SelfAttention($\cdot, \cdot, \cdot$)\\
Linear($D_{\text{feat}}$, $D_{\text{feat}}$)\\
Softmax()\\
\end{tabular}
& $D_{\text{weight}}$\\
\hline
(6) & (4)(5) &
\begin{tabular}[c]{@{}l@{}}
WeightedPooling()\\
Linear($D_{\text{feat}}$, 1) \end{tabular} & 
1 \\
\bottomrule
\end{tabular}
    }
\end{table*}

\bfpar{Training}
Our method is implemented in PyTorch and the 3D part is in PyTorch3D. We use  multi step learning rate with base learning rate of  $0.001$. The learning rate decays by a factor of $0.1$ after 100k and 200k iterations. We use the Stochastic Gradient Descent (SGD)
optimizer with a momentum of 0.9. The whole network is trained on a single NVIDIA A100 GPU with batch size 1 for 2 days. 

\section{Additional Results}
\label{sec:supp:results}
\subsection{Single camera frustums in Figure 3}

In \Fig{supp:fig3}, we show our predictions in green and purple cameras, extending \Fig{consistency}. SV-DRDF concatenates the single view predictions, so any inconsistency among the predictions affects the overall result.

\subsection{Additional Quantitative Results}
Extending~\Tab{tab1}, we include full evaluation metrics (Scene F score and Multiview Consistency) with different thresholds $\rho\in\{0.05, 0.1, 0.2, 0.5\}$. We also show accuracy and completeness for each method. 
We show single view in \Tab{supp:scene-pr-1view}; 3-view in \Tab{supp:scene-pr-3view}; 5-view in \Tab{supp:scene-pr-5view}. The results are consistent with \Tab{tab1} of the main paper across all thresholds. Depth-only oracle baseline have a slightly higher F score when evaluated with small $\rho$ because it has ground truth scale and shift. However, they cannot reconstruct any of them and thus get 0 for recall on hidden surfaces.

Our approach, designed for flexible input views, matches prior works in single-view scene reconstruction and achieves state-of-the-art results with multiple input views. In single-image cases, it is comparable to the single-view DRDF baseline. 

For 3-view sets, our method outperforms MCC~\cite{wu2023mcc} or DRDF~\cite{kulkarni2022directed}. MCC has higher recall in the low threshold regime but its prediction is blobby due to its occupancy prediction, thus having low precision and resulting in a low overall F score. 
SV-DRDF has a comparable score with us on visible surfaces, but is worse on hidden surfaces, resulting in a lower score for the full scene. 
When evaluated on scene consistency, our method shows a large absolute improvement over all baselines on all thresholds ($>11\%$ absolute improvement when $\rho=0.05, 0.1$), demonstrating that fusing features from multiple views enables a more coherent prediction.

The trend persists with 5-view inputs, where our method has the highest F score and consistency. Our method is not trained on 5-views subset but still remains robust to more input views enhancing the reconstruction quality in both visible and hidden surface reconstructions.

\begin{figure*}[t]
\centering
\resizebox{\textwidth}{!}{
\begin{tabular}{c@{\hskip4pt}c}
\includegraphics[width=0.3\textwidth]{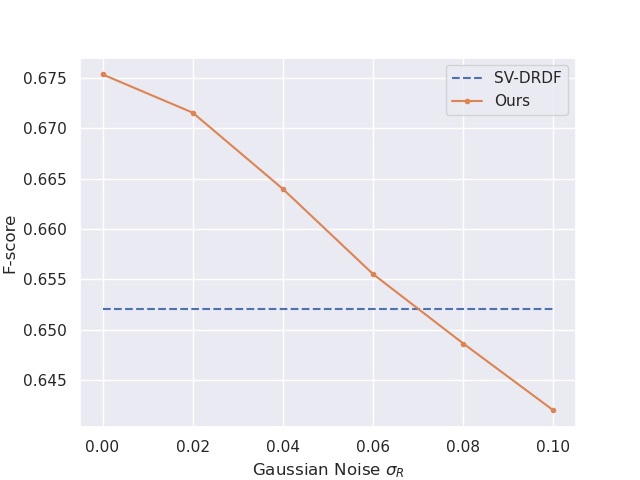} &
\includegraphics[width=0.3\textwidth]{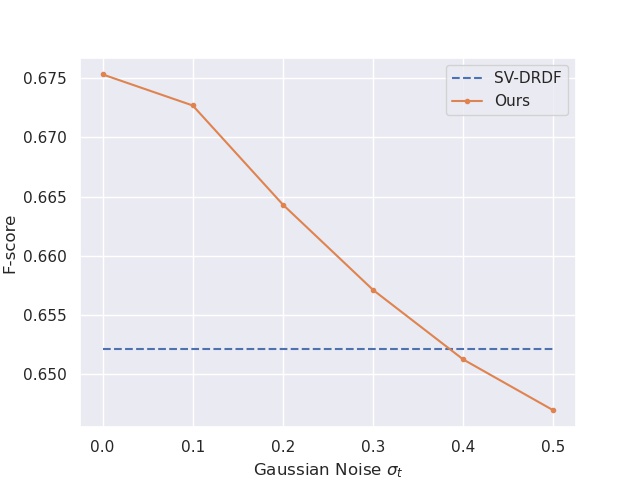} \\
\end{tabular}
}
    
    \caption{F score with respect to different noise level. Our method is robust to multi-view features with small noisy camera poses. It is better than single view DRDF baseline when $\sigma_R\leq0.06$, corresponding to 6.44$^\circ$ rotation error; or $\sigma_t\leq0.3$ corresponding to 0.56m translation error.}
    \label{fig:gaussian_camera}
\end{figure*}

\begin{table*}[h!]
\centering
\caption{Quantitative results for single view test scenes.
} %

\resizebox{\linewidth}{!}{ %
\begin{tabular}{@{}lcccccccccccc@{}}
\textit{Hidden}\\ 
\toprule
 & p@0.05 & p@0.1 & p@0.2 & p@0.5 & r@0.05 & r@0.1 & r@0.2 & r@0.5 & f@0.05 & f@0.1 & f@0.2 & f@0.5 \\
\hline
Depth only & - & - & - & - & 0 & 0 & 0 & 0 & - & - & - & - \\
\hline
MCC & 11.84 & 25.90 & 45.57 & 74.42 & 4.96 & 20.37 & 40.87 & 64.81 & 6.27 & 20.87 & 40.27 & 66.09 \\
SV-DRDF & \textbf{22.57} & \textbf{44.07} & \textbf{62.46} & \textbf{82.04} & 12.82 & 30.45 & 50.77 & 76.10 & 15.07 & 33.77 & \textbf{53.36} & 76.64 \\
Ours & 21.13 & 41.18 & 58.77 & 80.01 & \textbf{13.03} & \textbf{31.32} & \textbf{52.85} & \textbf{79.23} & \textbf{15.11} & \textbf{33.79} & 53.34 & \textbf{77.36} \\
\bottomrule
\\
\textit{Visible}\\
\toprule
 & p@0.05 & p@0.1 & p@0.2 & p@0.5 & r@0.05 & r@0.1 & r@0.2 & r@0.5 & f@0.05 & f@0.1 & f@0.2 & f@0.5 \\
\hline
Depth only & \textbf{46.27} & \textbf{71.41} & \textbf{84.59} & \textbf{93.20} & \textbf{36.94} & \textbf{66.63} & \textbf{87.39} & \textbf{97.35} & \textbf{40.14} & \textbf{68.05} & \textbf{85.31} & \textbf{94.96} \\
\hline
MCC & 14.00 & 30.76 & 55.67 & 86.51 & 10.38 & 37.91 & 59.67 & 76.45 & 11.25 & 32.87 & 56.40 & 80.44 \\
SV-DRDF & 35.42 & 60.35 & 78.45 & 92.20 & 25.25 & 49.56 & 71.15 & 89.38 & 28.44 & 52.99 & 73.45 & 90.03 \\
Ours & 35.95 & 61.44 & 79.04 & 92.36 & 25.53 & 50.48 & 72.10 & 90.20 & 28.75 & 54.05 & 74.29 & 90.57 \\
\bottomrule
\\
\textit{Full Scene}\\
\toprule
 & p@0.05 & p@0.1 & p@0.2 & p@0.5 & r@0.05 & r@0.1 & r@0.2 & r@0.5 & f@0.05 & f@0.1 & f@0.2 & f@0.5 \\
\hline
Depth only & \textbf{46.27} & \textbf{71.41} & \textbf{84.59} & \textbf{93.20} & \textbf{21.64} & 38.15 & 48.90 & 53.45 & \textbf{28.21} & \textbf{47.92} & 60.12 & 66.11 \\
\hline
MCC & 13.17 & 29.02 & 52.54 & 82.98 & 7.83 & 29.84 & 50.59 & 70.13 & 9.23 & 28.31 & 50.25 & 74.84 \\
SV-DRDF & 31.28 & 55.09 & 73.20 & 88.94 & 19.41 & 40.41 & 61.03 & 82.35 & 22.96 & 45.15 & 65.21 & 84.49 \\
Ours & 30.71 & 54.38 & 71.87 & 87.99 & 19.80 & \textbf{41.47} & \textbf{62.61} & \textbf{84.35} & 23.09 & 45.73 & \textbf{65.71} & \textbf{85.18} \\
\bottomrule
\end{tabular}
} %

\label{tab:supp:scene-pr-1view}
\end{table*}

\begin{table*}[t]
\centering
\caption{Quantitative results for 3-view test scenes.
} %

\resizebox{\linewidth}{!}{ %
\begin{tabular}{@{}lcccccccccccc@{}}
\textit{Hidden [41.9\%]}\\ 
\toprule
 & p@0.05 & p@0.1 & p@0.2 & p@0.5 & r@0.05 & r@0.1 & r@0.2 & r@0.5 & f@0.05 & f@0.1 & f@0.2 & f@0.5 \\
\hline
Depth only & - & - & - & - & 0 & 0 & 0 & 0 & - & - & - & - \\
\hline

MCC & 11.47 & 26.35 & 46.97 & 77.15 & \textbf{18.54} & \textbf{32.69} & 43.92 & 60.11 & 12.99 & 27.23 & 42.91 & 65.33 \\
SV-DRDF & 17.93 & 38.14 & 59.03 & 84.10 & 11.33 & 25.51 & 42.93 & 67.26 & 13.16 & 29.12 & 48.02 & 73.28 \\
Ours & \textbf{18.41} & \textbf{39.05} & \textbf{60.08} & \textbf{84.95} & 11.88 & 26.42 & \textbf{44.97} & \textbf{70.61} & \textbf{13.79} & \textbf{30.22} & \textbf{49.99} & \textbf{75.87} \\
\bottomrule
\\
\textit{Visible [58.1\%]}\\
\toprule
 & p@0.05 & p@0.1 & p@0.2 & p@0.5 & r@0.05 & r@0.1 & r@0.2 & r@0.5 & f@0.05 & f@0.1 & f@0.2 & f@0.5 \\
\hline
Depth only & \textbf{45.32} & \textbf{73.84} & \textbf{87.95} & \textbf{95.98} & \textbf{40.54} & \textbf{69.31} & \textbf{88.19} & \textbf{97.33} & \textbf{42.26} & \textbf{71.13} & \textbf{87.84} & \textbf{96.56} \\
\hline
MCC & 15.29 & 34.53 & 60.98 & 90.10 & 35.50 & 53.68 & 64.33 & 78.90 & 20.99 & 41.49 & 62.02 & 83.72 \\
SV-DRDF & 33.10 & 60.83 & 79.35 & 93.39 & 30.63 & 54.27 & 73.94 & 90.12 & 31.24 & 56.73 & 76.19 & 91.50 \\
Ours & 34.35 & 62.45 & 80.60 & 93.98 & 30.57 & 53.96 & 73.89 & 90.28 & 31.81 & 57.33 & 76.74 & 91.85 \\
\bottomrule
\\
\textit{Full Scene}\\
\toprule
 & p@0.05 & p@0.1 & p@0.2 & p@0.5 & r@0.05 & r@0.1 & r@0.2 & r@0.5 & f@0.05 & f@0.1 & f@0.2 & f@0.5 \\
\hline
Depth only & \textbf{45.32} & \textbf{73.84} & \textbf{87.95} & \textbf{95.98} & 24.50 & 41.38 & 52.01 & 56.85 & \textbf{30.69} & \textbf{51.55} & 63.90 & 69.99 \\
\hline
MCC & 14.28 & 32.37 & 57.43 & 86.91 & \textbf{27.46} & \textbf{43.75} & 54.43 & 69.64 & 18.26 & 36.28 & 54.78 & 76.39 \\
SV-DRDF & 30.02 & 56.18 & 75.15 & 91.44 & 22.16 & 41.26 & 59.48 & 79.08 & 24.79 & 46.61 & 65.61 & 84.25 \\
Ours & 30.81 & 57.21 & 75.94 & 91.91 & 22.26 & 41.42 & \textbf{60.33} & \textbf{80.72} & 25.27 & 47.24 & \textbf{66.56} & \textbf{85.47} \\
\bottomrule
\\
\textit{Ray Consistency}\\
\toprule
 & @0.05 & @0.1 & @0.2 & @0.5 \\
\hline
Depth only & 27.92 & 52.27 & 72.79 & 91.24 \\
\hline
MCC & 40.91 & 59.93 & 70.20 & 85.32 \\
SV-DRDF & 28.68 & 54.23 & 76.44 & 94.00 \\
Ours & \textbf{40.94} & \textbf{66.91} & \textbf{85.48} & \textbf{97.24} \\
\bottomrule
\end{tabular}
} %

\label{tab:supp:scene-pr-3view}
\end{table*}

\begin{table*}[t]
\centering
\caption{Quantitative results for 5-view test scenes.
} %

\resizebox{\linewidth}{!}{ %
\begin{tabular}{@{}lcccccccccccc@{}}
\textit{Hidden [43.7\%]}\\ 
\toprule
 & p@0.05 & p@0.1 & p@0.2 & p@0.5 & r@0.05 & r@0.1 & r@0.2 & r@0.5 & f@0.05 & f@0.1 & f@0.2 & f@0.5 \\
\hline
Depth only & - & - & - & - & 0 & 0 & 0 & 0 & - & - & - & - \\
\hline
MCC & 10.91 & 25.24 & 44.69 & 73.89 & 6.34 & 21.64 & 36.60 & 53.50 & 7.30 & 21.85 & 38.51 & 60.05 \\
SV-DRDF & 15.99 & 35.63 & 57.35 & 84.27 & 10.73 & 25.00 & 42.58 & 67.83 & 12.14 & 28.16 & 47.51 & 73.95 \\
Ours & \textbf{16.87} & \textbf{37.17} & \textbf{58.95} & \textbf{85.79} & \textbf{11.16} & \textbf{25.97} & \textbf{44.68} & \textbf{71.14} & \textbf{12.77} & \textbf{29.40} & \textbf{49.52} & \textbf{76.58} \\
\bottomrule
\\
\textit{Visible [56.3\%]}\\
\toprule
 & p@0.05 & p@0.1 & p@0.2 & p@0.5 & r@0.05 & r@0.1 & r@0.2 & r@0.5 & f@0.05 & f@0.1 & f@0.2 & f@0.5 \\
\hline
Depth only & \textbf{46.91} & \textbf{77.30} & \textbf{91.35} & \textbf{98.08} & \textbf{46.18} & \textbf{74.41} & \textbf{91.41} & \textbf{98.48} & \textbf{46.19} & \textbf{75.60} & \textbf{91.29} & \textbf{98.26} \\
\hline
MCC & 15.87 & 36.05 & 63.99 & 91.93 & 17.09 & 47.97 & 65.55 & 81.10 & 16.25 & 40.88 & 64.44 & 85.96 \\
SV-DRDF & 35.68 & 64.60 & 82.66 & 95.35 & 37.00 & 62.11 & 80.27 & 93.99 & 35.94 & 62.97 & 81.31 & 94.62 \\
Ours & 36.01 & 65.94 & 84.00 & 95.89 & 36.10 & 61.16 & 79.86 & 93.83 & 35.64 & 63.12 & 81.74 & 94.79 \\
\bottomrule
\\
\textit{Full Scene}\\
\toprule
 & p@0.05 & p@0.1 & p@0.2 & p@0.5 & r@0.05 & r@0.1 & r@0.2 & r@0.5 & f@0.05 & f@0.1 & f@0.2 & f@0.5 \\
\hline
Depth only & \textbf{46.91} & \textbf{77.30} & \textbf{91.35} & \textbf{98.08} & \textbf{29.53} & \textbf{47.26} & 57.62 & 61.79 & \textbf{35.30} & \textbf{57.33} & 69.40 & 74.65 \\
\hline
MCC & 14.94 & 34.03 & 60.47 & 88.66 & 12.68 & 37.12 & 53.35 & 69.36 & 13.40 & 34.86 & 55.94 & 77.16 \\
SV-DRDF & 32.75 & 60.36 & 78.95 & 93.79 & 26.51 & 47.22 & 64.74 & 82.75 & 28.71 & 52.21 & 70.54 & 87.58 \\
Ours & 33.00 & 61.42 & 80.00 & 94.28 & 25.99 & 46.91 & \textbf{65.28} & \textbf{84.03} & 28.49 & 52.47 & \textbf{71.41} & \textbf{88.57} \\
\bottomrule
\\
\textit{Ray Consistency}\\
\toprule
 & @0.05 & @0.1 & @0.2 & @0.5 \\
\hline
Depth only & 27.27 & 51.77 & 72.57 & 90.47 \\
\hline
MCC & 13.63 & 47.26 & 66.57 & 83.07 \\
SV-DRDF & 29.48 & 56.04 & 78.13 & 94.31 \\
Ours & \textbf{40.20} & \textbf{67.10} & \textbf{85.92} & \textbf{97.23} \\
\bottomrule
\end{tabular}
} %

\label{tab:supp:scene-pr-5view}
\end{table*}

\subsection{Robustness analysis with synthetic camera noise.}
How robust is our method if the camera pose is noisy? We test if the misaligned image features caused by noisy camera projection matrices can degrade the performance of our system. 
We keep one camera at the origin and add random noise to other camera projection matrices. We evaluate the reconstruction within the camera at the origin. Specifically, we parametrize the rotation $\mathfrak{so}(3)$ Lie algebra, similar to~\cite{lin2021barf} and synthetically perturb the camera poses with
additive noise $\delta \RB \sim \mathcal{N} (0, \sigma_R)$ and translation in $\delta \tB \sim \mathcal{N} (0, \sigma_t)$, where $\sigma_R$ in (0.02, 0.04, 0.06, 0.08, 0.10), corresponding to a 95 percentile error at (3.20, 6.37, 9.55, 12.82, 15.97) [$^\circ$] in rotation angle and $\sigma_t$ in (0.1, 0.2, 0.3, 0.4, 0.5), corresponding to a 95 percentile error at (0.28, 0.56, 0.84, 1.12, 1.40)[m] in translation. We evaluate the F score under different noise levels. 
Results are shown in \Fig{gaussian_camera}. Our method is robust to multi-view features with small noisy camera poses. It is better than single view DRDF baseline when $\sigma_R\leq0.06$, corresponding to 9.55$^\circ$ rotation error; or $\sigma_t\leq0.3$ corresponding to a big 0.84m translation error.

\subsection{More qualitative results}
Please check the video or the interactive demo at \url{https://jinlinyi.github.io/3DFIRES/}.

\end{document}